\documentclass[a4paper,11pt]{amsart}%

\usepackage{cite}

   \usepackage{graphicx}
    \usepackage{tikz}

\usepackage{amsmath}
\usepackage{amsfonts}
\usepackage{amssymb}
 \usepackage{multirow}

\usepackage{fullpage}

\usepackage{array}

  \usepackage[caption=false,font=normalsize,labelfont=sf,textfont=sf]{subfig}

\usepackage{stfloats}

\begin{document}

\title{A generalized multivariate Student-t mixture model for Bayesian classification and clustering of radar waveforms}

\author{Guillaume~Revillon,~Student Member,~IEEE,
        Ali~Mohammad-Djafari,~Member,~IEEE,
        and~Cyrille~Enderli} 
\thanks{M. Revillon is with Laboratoire des Signaux Syst\`{e}mes, Centrale Sup\'{e}lec, Universit\'{e} Paris Saclay and Thales Syst\`{e}mes A\'{e}roport\'{e}s, France. Email : guillaume.revillon@l2s.centralesupelec.fr}
\thanks{M. Mohammad-Djafari is with Laboratoire des Signaux Syst\`{e}mes, Centrale Sup\'{e}lec, Universit\'{e} Paris Saclay and CNRS, France. Email : Ali.mohammad-Djafari@l2s.centralesupelec.fr}
\thanks{M. Enderli is with Thales Syst\`{e}mes A\'{e}roport\'{e}s, France. Email : cyrille-jean.enderli@fr.thalesgroup.com}


\maketitle

\begin{abstract}
In this paper, a generalized multivariate Student-t mixture model is developed for classification and clustering of Low Probability of Intercept radar waveforms. A Low Probability of Intercept radar signal is characterized by a pulse compression waveform which is either frequency-modulated or phase-modulated. The proposed model can classify and cluster different modulation types such as linear frequency modulation, non linear frequency modulation, polyphase Barker, polyphase P1, P2, P3, P4, Frank and Zadoff codes. The classification method focuses on the introduction of a new  prior distribution for the model hyper-parameters that gives us the possibility to handle sensitivity of mixture models  to initialization and  to  allow  a less restrictive  modeling of data. Inference is processed through a Variational Bayes method and a Bayesian treatment is adopted for model learning, supervised classification and clustering. Moreover, the novel prior distribution is not a well-known probability distribution  and both deterministic and stochastic methods are employed to estimate  its expectations. Some numerical experiments show that the proposed method is less sensitive to initialization and  provides  more accurate results than the previous state of the art mixture models.
\end{abstract}

\smallskip
\noindent \textbf{Keywords.} Bayesian inference, generalized Student-t distribution, robust clustering

\section{Introduction}
%
%
%
%
In electronic warfare \cite{Schleher1986}, radar signal identification is a crucial component of Electronic Support Measures (ESM) systems. By providing information about the presence of threats, classification of radar signal has a self protection role  ensuring that countermeasures against enemy are well-chosen by  ESM systems \cite{Wiley1982}. Furthermore, improvement of the electronic intelligence database is a real challenge for military intelligence and clustering of radar signal can take a significant part in it by detecting unknown signal waveforms. Through its classification and clustering aspects, identification of radar signal is an important asset for decision making in military tactical situations. \\

To avoid identification of operating radars by ESM systems, radar designers have developed Low Probability of Intercept (LPI) waveforms. Theses waveforms are either frequency-modulated  or phased-modulated in order to improve resolution for the radar emitter at the expense of a suboptimal signal-to-noise ratio (SNR) \cite{Levanon2004}. In other words, theses pulse modulations allow to maximize the target range and the range resolution of radars. On the contrary, ESM resolution is less accurate since LPI signals are embedded in much noise and the identification task can be compromised. Most of classification approaches for intrapulse modulations are based on features extraction such as in \cite{Milne2002,Copeland2002} where Choi-Williams distribution, Wigner-Ville distribution and quadrature mirror filter bank time-frequency techniques are carried out. In \cite{Gencol2016}, statistical features are extracted of amplitude histograms and lead to a separation of common modulation types. Finally, \cite{Ravi2017} propose an automatic  and computationally less intensive feature extraction method based on Radon transform and fractional Fourier transform. However, these supervised approaches do not handle clustering issue and hence a mixture model approach is preferred.\\

Mixture modeling  \cite{Jordan1994} is a natural  framework for classification and  clustering. It can be formalized  as : 
        
        	\begin{equation}\label{mixture}
        		p(\mathbf{x}|\boldsymbol{\Theta},K)=\sum\limits_{k=1}^{K}a_k\psi_k(\mathbf{x}|{\boldsymbol{\theta}}_k)~,
        	\end{equation}

\begin{flushleft}
	where $\mathbf{x}\in\mathcal{X}\subseteq \mathbb{R}^d$ is an observation and $\boldsymbol{\Theta}=(\mathbf{a},\boldsymbol{\theta}_1,\ldots,\boldsymbol{\theta}_K)$, with $\mathbf{a}=[a_1,\ldots,a_K]'$, stands for parameters. Each probability distribution $\psi_k$ stands for the $k^{th}$  component distribution  with a weight $a_k$ where $a_k\geq 0$ and $\sum_ka_k=1$. 
\end{flushleft}

            Gaussian mixture models \cite{Quandt1978} (GMM)  have been widely used for decades. However, a major limitation of GMMs is their lack of robustness to outliers that can leads to over-estimate the number of clusters since they use additional components to capture the tails of the distributions \cite{Svensen2005}. Since LPI signals are embedded in noise, we propose to use a mixture of Student-t distributions in a Bayesian framework. The main advantages of this model is that the model accounts for the uncertainties of variances and covariances since the Student-t component is heavy-tailed \cite{Archambeau2007}. Student-t mixture models have performed well in many classification and clustering problems such  as features selection  \cite{Zhang2012,Sun2017} or image segmentation  \cite{Zhang2013,Nguyen2014}. Exact inference in that Bayesian approach is unfortunately intractable and a Variational Bayesian (VB) inference \cite{Waterhouse1996} is used to estimate the posterior distribution. The multivariate Student-t distribution is defined as follows
            
\begin{equation}\label{studentdistr}
	\mathcal{T}(\mathbf{x}|\boldsymbol{\mu}, \boldsymbol{\Sigma}, \nu)=c_{\mathcal{T}}(\nu,d)\times |\boldsymbol{\Sigma}^{-1}|^{\frac{1}{2}}
    \times  \left[1+\frac{1}{\nu}\mathcal{D}(\mathbf{x},\boldsymbol{\mu},\boldsymbol{\Sigma})\right]^{-\frac{d+\nu}{2}}
\end{equation}

 	\begin{flushleft}
 		where $d$ is the dimension of the feature space, $ \boldsymbol{\mu}$ and $\boldsymbol{\Sigma}$ are respectively the component mean and the component covariance matrix, $c_{\mathcal{T}}(\nu,d)=\frac{\Gamma(\frac{d+\nu}{2})}{\Gamma(\frac{\nu}{2})(\nu \pi)^{\frac{d}{2}}}$ is the normalizing constant and
 	\end{flushleft}
	
	\vspace{-10pt}
	
	\begin{equation*}
\mathcal{D}(\mathbf{x},\boldsymbol{\mu},\boldsymbol{\Sigma})=(\mathbf{x}-\boldsymbol{\mu})^T\boldsymbol{\Sigma}^{-1}(\mathbf{x}-\boldsymbol{\mu})\; .
\end{equation*}
    
 Assuming a dataset $\mathbf{X}\in \mathbb{R}^{d\times N}$ of i.i.d observations $(\mathbf{x}_1,\ldots,\mathbf{x}_N)$, a Student-t mixture is  then a weighted sum of multivariate Student-t distributions such that
 
	\begin{equation}\label{studentmixt}			p(\mathbf{X}|\boldsymbol{\Theta},K)=\prod\limits_{n=1}^N\sum\limits_{k=1}^K a_k\mathcal{T}(\mathbf{x_n}|\boldsymbol{\theta}_k),
     \end{equation}
          
         \begin{flushleft}
         	where $\boldsymbol{\theta}_k=\left(\nu_k,\boldsymbol{\mu}_k,\boldsymbol{\Sigma}_k\right)$.
         \end{flushleft} 
          
In former studies \cite{Svensen2005,Archambeau2007,Zhang2012,Sun2017}, $\nu_k$ has been considered as a deterministic variable updated via an optimization argument during the maximization step of the VB inference. However, this assumption can be restrictive since it requires an initialization value for $\nu$ that can lead optimization procedure to a local optima. Here, a novel hierarchical architecture is introduced by incorporating two sets of random variables $(\boldsymbol{\alpha},\boldsymbol{\beta})$ as more generalized parameters for (\ref{studentdistr}). A conjugate prior distribution for $(\boldsymbol{\alpha},\boldsymbol{\beta})$ is defined to avoid a non closed-form posterior distribution during the VB inference and since  its posterior expectations are intractable, both deterministic and stochastic approximation methods are deployed to estimate them. Finally, Student-t distributions centers $\boldsymbol{\mu}_k$ are used  to be initialized with  results of clustering algorithms  such in \cite{Zhang2012,Sun2017} where a K-means clustering algorithm is applied for initialization. In this paper, a non-supervised initialization is proposed and experiments on various data show that  the proposed algorithm is less sensitive to initialization than the standard algorithm.\\

The paper is organized as follows. The LPI signal framework is shortly presented in Section \ref{sec:data}. After introducing the standard Student-t model, the generalized model and the novel prior are explained in Section \ref{sec:model}. VB inference procedure is derived in Section \ref{sec:inference} to obtain posterior distribution of mixture parameters. In Section \ref{sec:expectations}, a combination of Laplace approximation and Importance sampling  is proposed to compute the intractable posterior expectations. Finally, performances of the proposed method are detailed in Section \ref{sec:experiments}.

%
%
%

\section{LPI Signal framework}
\label{sec:data}
Thanks to complex intrapulse modulations, a LPI radar signal is weaker than standard radar signals and is emitted over a wide frequency band. These properties make ESM systems less sensitive to LPI signals since ESM systems can interpret LPI signal as noise. 
A basic discrete time representation is considered for a pulse signal $s(t)$ of duration $T$  such 

\begin{equation}\label{eqsignal}
	s(t)=x(t)+n(t)~,
\end{equation}
where $n(t)$ is a Gaussian noise with zero mean and $\sigma^2_n$ variance and $x(t)$ is given by
\begin{equation}\label{pulse_equation}
	x(t)=\Re\left(u(t)\exp(2i\pi f_0t)\right)~,
\end{equation}
 where $u(t)$ is the complex envelope of $x(t)$ and $f_0$ the reference frequency of the pulse.\\

When $u(t)$ contains a phase modulation pattern, the signal $x(t)$ is phase-modulated. On the contrary, if a frequency modulation pattern is observed in $u(t)$, the signal $x(t)$ is frequency modulated. These two most common pulse compression techniques are presented in the following subsections. 
 
 \subsection{Frequency Modulation Signal}\label{FM}
By spreading energy over a modulation bandwidth, Frequency Modulation (FM) signals provide a better range resolution than constant frequency signals.\\

 \subsubsection{Linear FM signal}
 In a Linear FM (LFM) signal, the frequency band $f_0 \pm B$ is swept linearly during the pulse duration $T$ such that 
 
 \begin{equation}\label{chirp}
 u(t)=\frac{1}{\sqrt T}rect\left(\frac{t}{T}\right)\exp(i\pi kt^2),
 \end{equation}
  where $k=\pm \frac{B}{T}$.
 Combining (\ref{pulse_equation}) and (\ref{chirp}), the LFM signal is obtained. However, LFM signals exhibit relatively high autocorrelation sidelobes and some form of amplitude weighting is necessary to reduce the autocorrelation sidelobes.\\
 
  \subsubsection{Nonlinear FM Signal and Costas code} 
  Nonlinear FM (NLFM) signal has a spectrum shaped by deviating the constant rate of frequency change and by spending more time at frequencies that need to be enhanced. In this paper, Quadratic FM (QFM) is presented as a polynomial extension of the LFM signal and is obtained by :
  
  \begin{equation}\label{quadchirp}
 u(t)=\frac{1}{\sqrt T}rect\left(\frac{t}{T}\right)\exp(i\pi( kt^2+k_1t^3)),
 \end{equation}
  where 
  \begin{align*}
  &k=\pm \frac{B}{T}~,\\
  &k_1=\pm \frac{2B}{3T^2}~.
  \end{align*}
  
  The Costas code \cite{Costas1984} results in a rather randomlike frequency evolution on a band B. $M$ distinct frequencies, equally spaced by $\frac{B}{M}$, are only transmitted once on one of  $M$ equal time slices of duration $t_b=\frac{T}{M}$. \cite{Levanon2004} gives the following definition of the complex envelope of a Costas signal 
 
 \begin{equation}\label{costas}
 	 u(t)=\frac{1}{\sqrt t_b}\sum_{m=1}^Mu_m(t-(m-1)t_b)~,
 \end{equation}
 where

\begin{equation*}
  u_m(t)=
     \begin{cases}
        \exp(2i\pi \frac{f^c_m}{t_b}t), & 0\le t\le t_b, \\
        0  & \text{elsewhere}~,
     \end{cases}
\end{equation*}

\begin{flushleft}
with $f^c=[f^c_1,f^c_2,\ldots,f^c_M]$ is the Costas frequencies sequence.  Combining both (\ref{quadchirp}) and (\ref{costas}) with  (\ref{pulse_equation}), NLFM and Costas signals are obtained.
 \end{flushleft}
 
 \subsection{Phase Modulation Signal}\label{PM}
 Phase coding is one of the first methods for pulse compression. The concept rests on dividing a pulse of duration $T$ into $M$ bits of identical duration $t_b=\frac{T}{M}$ and assigning a different phase value to each bit.
 The main advantage of phase coding over frequency modulation is low peak side lobe level  \cite{Levanon2004}. The complex envelope of each proposed phase code is given by 
 \begin{equation}
 	u(t)=\frac{1}{\sqrt T}\sum_{m=1}^Mu_mrect\left[ \frac{t-(m-1)t_b}{t_b}\right]~,
 \end{equation}
 where $u_m=\exp(i\psi_m)$ and $\psi=[\psi_1,\psi_2,\ldots,\psi_M]$ is the phase code.\\
 
  \subsubsection{Barker codes}
  Barker codes were introduced by  \cite{Barker1953} and is one of the most famous family of phase codes. All known binary sequences yielding a peak-to-peak sidelobe ratio of $M$ were reported by  \cite{Barker1953} and \cite{Turyn1963}  and are given in Table \ref{barker}. \\

  \begin{table}[]
\centering
\caption{All known binary Barker codes }
\label{barker}
\begin{tabular}{cc}
\hline
Code length & Code          \\ \hline
2           & 11 or 10      \\
3           & 110           \\
4           & 1110 or 1101  \\
5           & 11101         \\
7           & 1110010       \\
11          & 11100010010   \\
13          & 1111100110101 \\ \hline
\end{tabular}
\end{table}

 \subsubsection{Frank, P1 and P2 codes}
 
 The Frank code \cite{Frank1962} is a polyphasecode with a perfect square length ($M=L^2$).  The $M$-element Frank code is formed by concatenating the rows  of a $L\times L$ matrix $\psi$ whose elements are given by
 
 \begin{equation*}
 \psi_{i,j}=\frac{2\pi}{L}(i-1)(j-1)\;,
 \end{equation*}
 where $i=(1,\ldots,L)$ and $j=(1,\ldots,L)\;.$\\
 
 P1 and P2 codes are modified versions of Frank code and are given by 
 \begin{equation*}
\text{P1 code : }  \psi_{i,j}=-\frac{\pi}{L}(L-2j-1)((j-1)L+(i-1))
 \end{equation*}
 \begin{flushleft}
and
\end{flushleft}
 \begin{equation*}
 \text{P2 code : } \psi_{i,j}=\left(\frac{\pi}{2}\frac{L-1}{L}-\frac{\pi}{L}(i-1)\right)(L+1-2j)\;.
 \end{equation*}
 
  \subsubsection{Zadoff, P3 and P4 codes}
  
  While the Frank, P1 and P2 codes are only applicable for perfect square lengths, the Zadoff code \cite{Zadoff1963} is applicable for any length and is given by :
  \begin{equation*}
  	\psi_m=\frac{2\pi}{M}(m-1)\left(r\frac{M-1-m}{2}-q\right)
  \end{equation*}
    where $1\le m \le M$, $0\le q\le M$ is any integer and $r$ is any integer relatively prime to $M$.\\
    
  P3 and P4 codes are specific cyclically shifted and decimated versions of the Zadoff code and are given by 
  
 \begin{equation*}
 \text{P3 code : } \psi_{m}=\frac{\pi}{2}(m-1)^2
 \end{equation*}
 
  \begin{flushleft}
and
\end{flushleft}
  
 \begin{equation*}
 \text{P4 code : } \psi_{m}=\frac{\pi}{2}(m-1)^2-\pi(m-1)
 \end{equation*}
  where  $1\le m \le M$ is any integer.

\section{Model}
\label{sec:model}
In this section,  the standard  Student-t mixture model (SMM) is presented  as a hierarchical latent variable model  before introducing the  proposed generalized Student-t mixture model (GSMM) and a new prior distribution. 

\subsection{ Standard Student-t mixture model}

  A Student-t mixture can be formalized as a latent model since the component label associated to each data point is unobserved. To this end, a discrete variable $\mathbf{Z}=\{ \mathbf{z_n}\}_{n=1}^{N}$, with $z_{nk}\in\{ 0,1\}$ such that $\sum\limits_{k=1}^Kz_{nk}=1,\forall n$, is introduced to indicate  which cluster the data $x_n$ belongs to. 
 Moreover, noting that the Student-t distribution (\ref{studentdistr}) can be written as the marginal of a Gaussian-Gamma Distribution 
          
          \begin{equation}\label{student}
          \mathcal{T}(\mathbf{x}|\boldsymbol{\mu}, \boldsymbol{\Sigma}, \nu)=\int_{0}^{+\infty} \mathcal{N}\left(\mathbf{x}|\boldsymbol{\mu},u^{-1}\boldsymbol{\Sigma}\right)\mathcal{G}\left( u|\frac{\nu}{2},\frac{\nu}{2}\right)du\;,
          \end{equation}
          
          \begin{flushleft}
          another latent variable  $\mathbf{U}=\{ \mathbf{u_n}\}_{n=1}^{N}$ is introduced such that  $u_{nk}\sim \mathcal{G}(\frac{\nu_k}{2},\frac{\nu_k}{2})$ where 
          \end{flushleft}
          
          \vspace{-10pt}
          
          \begin{equation*}
          \mathcal{G}(u|a,b)=\frac{b^a}{\Gamma(a)}x^{a-1}\exp(-bu)\;.
          \end{equation*}
          
          \begin{flushleft}
          Therefore a distribution of $p(\mathbf{X}|\mathbf{Z},\mathbf{U},\boldsymbol{\Theta},K)$ is obtained  :
          \end{flushleft}
          
          \vspace{-10pt}
          
          \begin{equation}
 p(\mathbf{X}|\mathbf{Z},\mathbf{U},\boldsymbol{\Theta},K)=\prod\limits_{k=1}^K\prod\limits_{n=1}^N\mathcal{N}(\mathbf{x}_n|\boldsymbol{\mu}_k,u_{nk}^{-1}\boldsymbol{\Sigma}_k)^{z_{nk}}~.
          \end{equation}
          
          Conjugate priors over $\mathbf{Z}$ and $\mathbf{U}$ are added to complete the hierarchical latent variable model,
          
      \begin{align*}
      &p(\mathbf{U}|\mathbf{Z},\boldsymbol{\Theta},K)=\prod\limits_{k=1}^K\prod\limits_{n=1}^N\mathcal{G}\left(u_{n,k}|\frac{\nu_k}{2},\frac{\nu_k}{2}\right)^{z_{nk}}~,\\
      &p(\mathbf{Z}|\mathbf{\Theta},K)=\prod\limits_{k=1}^K\prod\limits_{n=1}^Na_k^{z_{nk}}~.
      \end{align*}
      
      At last, the Bayesian framework imposes to specify priors for the parameters $\boldsymbol{\Theta}$. The resulting conjugate priors are 
      
      \begin{equation*}
          \left \{
          \begin{aligned}
          	&p(\mathbf{a}|K)=\mathcal{D}(\mathbf{a}|\kappa_0) \\
          	&p(\boldsymbol{\mu}|\boldsymbol{\Sigma},K) =\prod_{k=1}^K \mathcal{N}(\boldsymbol{\mu}_k|\boldsymbol{\mu}_0,\eta_0^{-1}\boldsymbol{\Sigma}_k) \\
          &p(\boldsymbol{\Sigma}|K) = \prod_{k=1}^K\mathcal{IW}(\boldsymbol{\Sigma}_k|\gamma_0,\boldsymbol{\Sigma}_0)~.
          \end{aligned}
          \right .
      \end{equation*}
      
         \begin{flushleft}
         where  the Dirichlet and the Inverse Wishart distributions are defined  as follows : 
         \end{flushleft}
         
         \vspace{-10pt}
         
         \begin{align*}
         &\mathcal{D}(\mathbf{a}|\boldsymbol{\kappa})=c_{\mathcal{D}}(\boldsymbol{\kappa})\prod\limits_{k=1}^Ka_k^{\kappa_k-1}~,\\
            &\mathcal{IW}(\boldsymbol{\Sigma}|\gamma,\mathbf{S})=c_{\mathcal{IW}}(\gamma,\mathbf{S})|\boldsymbol{\Sigma}|^{-\frac{\gamma+d+1}{2}}\exp\left(-\frac{1}{2}tr(\mathbf{S}\boldsymbol{\Sigma}^{-1})\right)\;,
         \end{align*}
         
         \begin{flushleft}
         where  $c_{\mathcal{D}}(\boldsymbol{\kappa})$ and $c_{\mathcal{IW}}(\gamma,\mathbf{S})$ are normalizing constants such that
         \end{flushleft}
         
         \vspace{-10pt}
         
         \begin{equation*}
         c_{\mathcal{D}}(\boldsymbol{\kappa})=\frac{\Gamma\left(\sum_{k=1}^K\kappa_k\right)}{\prod_{k=1}^K\Gamma(\kappa_k)},~c_{\mathcal{IW}}(\gamma,\mathbf{S})=\frac{|\mathbf{S}|^{\frac{\gamma}{2}}}{2^{\frac{d\gamma}{2}}\Gamma_d(\frac{\gamma}{2})}~.
         \end{equation*}
         
\subsection{Generalized model}

 The degree of freedom variable $\nu$ has been considered as a deterministic variable updated via an optimization argument during the maximization step of the VB inference \cite{Peel2000}. Indeed, \cite{Svensen2005,Archambeau2007,Sun2017,Nguyen2014} did not assume any prior distribution for $\nu$ since there do not exist any  known conjugate priors for $\nu$. However, this assumption can be restrictive since it requires an initialization value for $\nu$ that can lead optimization procedure to a local optima.
 Therefore, a novel  hierarchical architecture  is proposed by incorporating  positive random variables $(\boldsymbol{\alpha},\boldsymbol{\beta})$ as parameters for  (\ref{studentdistr}) such that a generalized  Student-t distribution is defined as  
 
 \begin{equation}\label{generalizedstudent}
 \begin{split}
 \mathcal{T}(\mathbf{x}|\boldsymbol{\mu}, \boldsymbol{\Sigma}, \alpha,\beta)=&c_{T}(\alpha,\beta,d)\times|\boldsymbol{\Sigma}^{-1}|^{\frac{1}{2}}\\
    \times &\left[1+\frac{1}{2\beta}\mathcal{D}(\mathbf{x},\boldsymbol{\mu},\boldsymbol{\Sigma})\right]^{-(\alpha+\frac{d}{2})}
 \end{split}
\end{equation}

\begin{flushleft}
with the normalizing constant $c_{T}(\alpha,\beta,d)=\frac{\Gamma(\alpha+\frac{d}{2})}{\Gamma(\alpha)(2\beta \pi)^{\frac{d}{2}}}$ .
\end{flushleft}

This new parametrization of (\ref{studentdistr}) induces a  generalized  mixture model derived from (\ref{studentmixt}) where

\begin{align*}     &p(\mathbf{U}|\mathbf{Z},\boldsymbol{\alpha},\boldsymbol{\beta},K)=\prod\limits_{k=1}^K\prod\limits_{n=1}^N\mathcal{G}\left(u_{n,k}|\alpha_k,\beta_k\right)^{z_{nk}}\;,\\
&p(\boldsymbol{\alpha},\boldsymbol{\beta}|K)=\prod\limits_{k=1}^Kp(\alpha_k,\beta_k)\;,
\end{align*}
\begin{figure}
\centering
		\begin{tikzpicture}[scale=0.6]
			\draw (-1,-1) rectangle (8,3.5);
			\draw (3,-3) rectangle (7,1);
			\draw (7.5,-0.5) node{$K$};
			\draw (6.5,-2.5) node{$N$};
			\draw (-2.5,-1) rectangle (-1.5,1);
			\draw[->](-1.5,0)--(-0.5,0);
			\draw (-2,0.5) node{$\gamma_0$};
			\draw (-2,-0.5) node{$\boldsymbol{\Sigma}_0$};
			\draw (-2.5,1.5) rectangle (-1.5,3.5);
			\draw[->](-1.5,2)--(1.5,0.2);
			\draw (-2,3) node{$\boldsymbol{\mu}_0$};
			\draw (-2,2) node{$\eta_0$};
			\draw (0,0) circle (0.5) ;
			\draw (0,0) node{$\boldsymbol{\Sigma}_k$};
			\draw [->] (0.5,0) -- (1.5,0);
			\draw [->] (0,-0.5) -- (4.5,-2.1);
			\draw (2,0) circle (0.5) ;
			\draw (2,0) node{$\boldsymbol{\mu}_k$};
			\draw [->] (2,-0.5) -- (4.5,-1.9);
			\draw (3,2) circle (0.5) ;
			\draw (3,2) node{$\alpha_{k}$};
			\draw [->] (3,1.5) -- (3.9,0.5);
			\draw (5,2) circle (0.5) ;
			\draw (5,2) node{$\beta_{k}$};
			\draw [->] (5,1.5) -- (4.1,0.5);
			\draw (4,0) circle (0.5) ;
			\draw (4,0) node{$u_{nk}$};
			\draw [->] (5.5,0) -- (4.5,0);
			\draw [->] (4,-0.5) -- (4.9,-1.5);
			\draw (6,0) circle (0.5) ;
			\draw (5,-2) node{$\mathbf{x}_{n}$};
			\draw (5,-2) circle (0.5) ;
			\draw (6,0) node{$z_{nk}$};
			\draw [->] (8.5,0) -- (6.5,0);
			\draw [->] (6,-0.5) -- (5.1,-1.5);
			\draw (9,0) circle (0.5) ;
			\draw (9,0) node{$\mathbf{a}$};
			\draw (11,0) circle (0.5) ;
			\draw (11,0) node{$\kappa_{0}$};
			\draw [->] (10.5,0) -- (9.5,0);
			\draw (1.5,4) rectangle (3.5,5);
			\draw (2,4.5) node{$p_0$};
			\draw (3,4.5) node{$r_0$};
			\draw [->] (2.5,4) -- (3,2.5);
			\draw (4.5,4) rectangle (6.5,5);
			\draw (5,4.5) node{$q_0$};
			\draw (6,4.5) node{$s_0$};
			\draw [->] (5.5,4) -- (5,2.5);
	\end{tikzpicture}
	
\caption{ Graphical representation of the generalized Student-t mixture model. The arrows represent conditional dependencies between the random variables. The K-plate represents the K mixture components and the N-plate the independent identically distributed observations $x_n$. Note that the scale variables $u_{nk}$ and the indicator variables $z_{nk}$ belong to both plates, indicating that there is one such variable for each mixture component and each observation.}
\label{graphical}
\end{figure}
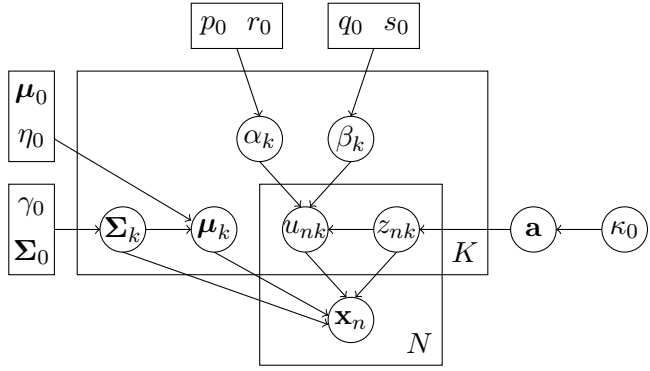
   To avoid a non closed-form posterior distribution for  each $(\alpha_k,\beta_k)$, a conjugate prior has to be chosen. Assuming that $(\alpha_k,\beta_k)$ are independent, the following prior is  introduced
   
    	\begin{equation}\label{indprior}
 	p(\alpha_k,\beta_k)\propto\frac{p_0^{\alpha_k-1}\beta_k^{s_0-1}e^{-q_0\beta_k}}{\Gamma(\alpha_k)^{r_0}}\mathbb{I}_{\{\alpha_k,\beta_k>0\}}\;,
    \end{equation}
    
    \begin{flushleft}
    where $p_0,q_0,r_0,s_0>0$.
    \end{flushleft}
    
     Marginalized priors  are then obtained
     
     \begin{align*}
    	&p(\beta_k)=\mathcal{G}(\beta_k|s_0,q_0)\;,\\
        &p(\alpha_k)=\frac{1}{M_0}\frac{p_0^{\alpha_k-1}}{\Gamma(\alpha_k)^{r_0}}\mathbb{I}_{\{\alpha_k>0\}}\;,
    \end{align*}
     
     \begin{flushleft}
     	where
     \end{flushleft}
     
      \vspace{-10pt}
      
     \begin{equation*}
    	M_0=\int\frac{p_0^{\alpha_k-1}}{\Gamma(\alpha_k)^{r_0}}\mathbb{I}_{\{\alpha_k>0\}}d\alpha_k\; .
    \end{equation*}
    Choosing $p_0\le 1$ ensures that $M_0$ is proper (Appendix A). The directed acyclic graph of the GSMM is shown in Figure \ref{graphical}.
    
\section{INFERENCE}
\label{sec:inference}
In this section, a brief introduction to Variational Bayes is proposed before developing  calculations of variational posterior  distributions related to latent variables $(\mathbf{U,Z})$ and parameters $(\boldsymbol{\Theta},\boldsymbol{\alpha},\boldsymbol{\beta})$. 

\subsection{Introduction to Variational Bayes } 

VB can be viewed as a Bayesian generalization of the Expectation-Maximization (EM) algorithm \cite{Dempster1977} combined with a Mean Field Approach \cite{Opper2001}. It consists in approximating the intractable posterior distribution $P=p(\mathbf{Z},\mathbf{U},\boldsymbol{\Theta},\boldsymbol{\alpha},\boldsymbol{\beta}|\mathbf{X},K)$ by a tractable one $Q=q(\mathbf{Z},\mathbf{U},\boldsymbol{\Theta},\boldsymbol{\alpha},\boldsymbol{\beta})$ whose parameters are chosen via a variational principle to minimize the Kullback-Leibler (KL) divergence 

\begin{equation*}
	KL\left[Q||P\right]=\int q(\mathbf{H}) \log\left(\frac{q(\mathbf{H})}{p(\mathbf{H}|\mathbf{X},K)}\right)d\mathbf{H}~,
\end{equation*}

\begin{flushleft}
   where $\mathbf{H}=\left(\mathbf{U},\mathbf{Z},\boldsymbol{\Theta},\boldsymbol{\alpha},\boldsymbol{\beta}\right)$.
 \end{flushleft}
Noting that $p(\mathbf{H}|\mathbf{X},K)=\frac{p(\mathbf{X},\mathbf{H}|K)}{p(\mathbf{X}|K)}$, the KL divergence can be written as 

\begin{equation*}
	KL\left[Q||P\right]=\log p(\mathbf{X}|K)-\mathcal{L}(Q)~.
\end{equation*}

$\mathcal{L}(Q)$ is considered as a lower bound for  the log evidence $\log p(\mathbf{X}|K)$ and can be expressed as

\begin{equation}\label{bound}
	\mathcal{L}(Q)=\mathbb{E}_q\left[\log p(\mathbf{X},\mathbf{H}|K)\right]-\mathbb{E}_q\left[\log q(\mathbf{H})\right]\;,
\end{equation}

\begin{flushleft}
	where $\mathbb{E}_q[\cdot]$ denotes the expectation with respect to $q$.
\end{flushleft}

Then, minimizing the KL divergence is equivalent to maximizing  $\mathbb{E}_q\left[\log p(\mathbf{X},\mathbf{H}|K)\right]$. Assuming that $q(\mathbf{H})$ can be factorized over the latent variables $(\mathbf{U},\mathbf{Z})$ and the  parameters $(\boldsymbol{\Theta},\boldsymbol{\alpha},\boldsymbol{\beta})$, a free-form maximization  with respect to  $q(\mathbf{U},\mathbf{Z})$, $q(\boldsymbol{\Theta})$, $q(\boldsymbol{\alpha})$ and $q(\boldsymbol{\beta})$ leads to  the following update rules  :

\begin{align*}
&\textbf{E-step} : q(\mathbf{U},\mathbf{Z})\propto \exp\left(\mathbb{E}_{\boldsymbol{\Theta},\boldsymbol{\alpha},\boldsymbol{\beta}}\left[\log p(\mathbf{X},\mathbf{U},\mathbf{Z}|\boldsymbol{\Theta},\boldsymbol{\alpha},\boldsymbol{\beta},K)\right]\right),\\
&\textbf{M-step} :  q(\boldsymbol{\Theta})\propto \exp\left(\mathbb{E}_{\mathbf{U},\mathbf{Z}}\left[\log p(\boldsymbol{\Theta}|\mathbf{X},\mathbf{Z},\mathbf{U},K)\right]\right),\\
&\boldsymbol{\alpha}\textbf{-step} :  q(\boldsymbol{\alpha})\propto \exp\left(\mathbb{E}_{\mathbf{U},\mathbf{Z},\boldsymbol{\beta}}\left[\log p(\boldsymbol{\alpha}|\mathbf{X},\mathbf{Z},\mathbf{U},K)\right]\right),\\
&\boldsymbol{\beta}\textbf{-step} :  q(\boldsymbol{\beta})\propto \exp\left(\mathbb{E}_{\mathbf{U},\mathbf{Z},\boldsymbol{\alpha}}\left[\log p(\boldsymbol{\beta}|\mathbf{X},\mathbf{Z},\mathbf{U},K)\right]\right)~.
\end{align*}

The expectations  $\mathbb{E}_{\mathbf{U},\mathbf{Z}}[\cdot]$, $\mathbb{E}_{\boldsymbol{\Theta}}[\cdot]$, $\mathbb{E}_{\boldsymbol{\alpha}}[\cdot]$ and $\mathbb{E}_{\boldsymbol{\beta}}[\cdot]$  are respectively taken with respect to the variational  posteriors $q(\mathbf{U},\mathbf{Z})$, $q(\boldsymbol{\Theta})$, $q(\boldsymbol{\alpha})$ and $q(\boldsymbol{\beta})$. Thereafter, the algorithm  iteratively updates the variational posteriors  by increasing  the bound  $\mathcal{L}(Q)$. Running the algorithm steps,  each posterior distribution is obtained in the following  subsections.

\subsection{Variational posterior distributions for latent variables}

Noting that $p(\mathbf{X},\mathbf{Z},\mathbf{U}|\boldsymbol{\Theta},\boldsymbol{\alpha},\boldsymbol{\beta})$ can be factorized as \\
$p(\mathbf{X}|\mathbf{U},\mathbf{Z},\boldsymbol{\Theta},\boldsymbol{\alpha},\boldsymbol{\beta})p(\mathbf{U}|\mathbf{Z},\boldsymbol{\Theta},\boldsymbol{\alpha},\boldsymbol{\beta})p(\mathbf{Z}|\boldsymbol{\Theta},\boldsymbol{\alpha},\boldsymbol{\beta})$, a factorized form $q(\mathbf{U}|\mathbf{Z})q(\mathbf{Z})$ is similarly chosen for $q(\mathbf{U},\mathbf{Z})$.
The E-step can be computed by developing the expectation 

\begin{equation}\label{expectation}
 				\begin{split}
 					&\mathbb{E}_{\boldsymbol{\Theta},\boldsymbol{\alpha},\boldsymbol{\beta}}\left[\log p(\mathbf{X},\mathbf{U},\mathbf{Z}|\boldsymbol{\Theta},\boldsymbol{\alpha},\boldsymbol{\beta},K)\right]=\sum_{k=1}^K\sum_{n=1}^N z_{nk}\bigg\{ \\
                    &-\frac{\mathbb{E}_{\boldsymbol{\Theta}}[\log \boldsymbol{|\Sigma}_k|]}{2}-\frac{d}{2}(\log 2\pi-\log u_{n,k}) + \mathbb{E}_{\boldsymbol{\Theta}}[\log a_k]\\
                    &-\frac{u_{n,k}}{2}\mathbb{E}_{\boldsymbol{\Theta}}\left[\mathcal{D}(\mathbf{x}_n,\boldsymbol{\mu}_k,\boldsymbol{\Sigma}_k)\right] + \mathbb{E}_{\boldsymbol{\alpha}}[\alpha_k]\mathbb{E}_{\boldsymbol{\beta}}[\log \beta_k]\\
                    &-\mathbb{E}_{\boldsymbol{\alpha}}[\log \Gamma (\alpha_k)]+ \left(\mathbb{E}_{\boldsymbol{\alpha}}[\alpha_k]-1\right)\log u_{n,k}-\mathbb{E}_{\boldsymbol{\beta}}[\beta_{k}]u_{n,k}\bigg\}\;.
  				\end{split}
			 \end{equation}
			
\begin{flushleft}
A conditional posterior  $q(\mathbf{U}|\mathbf{Z})$ is deduced from (\ref{expectation}) such that
\end{flushleft}

\begin{equation*}
	q(\mathbf{U}|\mathbf{Z})=\prod_{n=1}^N\prod_{k=1}^Kq(u_{nk}|z_{nk})\;,
\end{equation*}

\begin{flushleft}
where conditionally to each $z_{nk}\in \mathbf{Z}$
\end{flushleft}
        
 			\begin{equation*}
			  q(u_{nk}|z_{nk}=1)\sim 	\mathcal{G}(\tilde{\alpha}_{nk},\tilde{\beta}_{nk})
			 \end{equation*}
			  
			  \begin{flushleft}
			  where
			  \end{flushleft}
			  
			            \vspace{-12pt}

			  \begin{align*}
			  	&\tilde{\alpha}_{nk}=\mathbb{E}_{\boldsymbol{\alpha}}[\alpha_k]+\frac{d}{2}\;,\\
				&\tilde{\beta}_{nk}=\frac{1}{2}\mathbb{E}_{\boldsymbol{\Theta}}\left[\mathcal{D}(\mathbf{x}_n,\boldsymbol{\mu}_k,\boldsymbol{\Sigma}_k)\right]+\mathbb{E}_{\boldsymbol{\beta}}[\beta_k] \;.
			  \end{align*}
 Expectations of $\mathbf{U}$ are derived from the Gamma distribution properties such that 
 \begin{align*}
&\mathbb{E}_{\mathbf{U},\mathbf{Z}}[u_{nk}]=\frac{\tilde{\alpha}_{nk}}{\tilde{\beta}_{nk}}~,\\
 &\mathbb{E}_{\mathbf{U},\mathbf{Z}}[\log u_{nk}]=\psi(\tilde{\alpha}_{nk})-\log \tilde{\beta}_{nk}~,
 \end{align*}
\begin{flushleft}
where $\psi(\cdot)$ is the digamma function.
\end{flushleft}
Due to the conjugacy property, a conjugate posterior distribution for latent variable $\mathbf{Z}$ is obtained from  (\ref{expectation})

\begin{equation}\label{qz}
q(\mathbf{Z})=\prod_{n=1}^N\prod_{k=1}^K\rho_{nk}^{z_{nk}}\;,
\end{equation}

\begin{flushleft}
where $\rho_{nk}=q(z_{nk}=1)$ is called the responsibility.
\end{flushleft}

Instead of assuming that most of the probability mass of the posterior distribution of  the scale variable $\mathbf{U}$ is located around its mean (\cite{Svensen2005,Sun2017}), \cite{Archambeau2007} proposed to integrate out $\mathbf{U}$ the joint  variational posterior $q(\mathbf{U},\mathbf{Z})$ to obtain $q(\mathbf{Z})$. Therefore, it consists in substituting (\ref{expectation}) in the E-step and marginalizing  over $\mathbf{U}$.  That approach leads to the following responsibilities 

\begin{equation*}
	\begin{split}
		\rho_{nk} &\propto \int_0^\infty \exp\mathbb{E}_{\boldsymbol{\Theta},\boldsymbol{\alpha},\boldsymbol{\beta}}\left[\log p(x_n,u_{n,k},z_{n,k}|\boldsymbol{\Theta},\boldsymbol{\alpha},\boldsymbol{\beta},K)\right] du_{nk}\\
        &\propto \frac{\exp \left(\mathbb{E}_{\boldsymbol{\Theta}}[\log a_k]+\mathbb{E}_{\boldsymbol{\beta}}[\log \beta_k]\mathbb{E}_{\boldsymbol{\alpha}}[\alpha_k] \right)\Gamma\left(\mathbb{E}_{\boldsymbol{\alpha}}[\alpha_k]+\frac{d}{2}\right)}{\exp \left(\frac{\mathbb{E}_{\boldsymbol{\Theta}}[\log|\boldsymbol{\Sigma}_k|]}{2}+\mathbb{E}_{\boldsymbol{\alpha}}[\log\Gamma(\alpha_k)]\right) \mathbb{E}_{\boldsymbol{\beta}}[\beta_k]^{\left(\mathbb{E}_{\boldsymbol{\alpha}}[\alpha_k]+\frac{d}{2}\right)}}\\
        &\times \left[1+\frac{\mathbb{E}_{\boldsymbol{\Theta}}\left[\mathcal{D}(\mathbf{x}_n,\boldsymbol{\mu}_k,\boldsymbol{\Sigma}_k)\right]}{2\mathbb{E}_{\boldsymbol{\beta}}[\beta_k]}\right]^{-\left(\mathbb{E}_{\boldsymbol{\alpha}}[\alpha_k]+\frac{d}{2}\right)} \; .
	\end{split}
\end{equation*}


Then, the responsibilities are normalized as follows 

\begin{equation}\label{norm}
	r_{nk}=\frac{\rho_{nk}}{\sum_{k=1}^K\rho_{nk}}\;.
\end{equation}

Expectation of $\mathbf{Z}$ is deduced from (\ref{qz}) and is given by 
\begin{equation*}
\mathbb{E}_{\boldsymbol{U,Z}}[z_{n,k}]=r_{nk}\; .
\end{equation*}
          \subsection{Variational posterior distributions for parameters}
	Since $p(\boldsymbol{\Theta}|\mathbf{X},\mathbf{Z},\mathbf{U},K)$ can be decomposed as $p(\boldsymbol{a}|\mathbf{X},\mathbf{Z},\mathbf{U},K)p(\boldsymbol{\mu}|\boldsymbol{\Sigma},\mathbf{X},\mathbf{Z},\mathbf{U},K)p(\boldsymbol{\Sigma}|\mathbf{X},\mathbf{Z},\mathbf{U},K)$, the following similar form is chosen for $q(\boldsymbol{\Theta})$

         \begin{equation*}
         q(\boldsymbol{\Theta})=q(\boldsymbol{a})\prod_{k=1}^Kq(\boldsymbol{\mu}_k|\boldsymbol{\Sigma}_k)q(\boldsymbol{\Sigma}_k)~,
          \end{equation*}
        \begin{flushleft}
        Due to the conjugacy property, a conjugate distribution is obtained for $\boldsymbol{\Theta}$ 
	\end{flushleft}
           \begin{equation*}
          \left \{
          \begin{aligned}
          	&q(\mathbf{a}|K)=\mathcal{D}(\boldsymbol{a}|\tilde{\boldsymbol{k}})~, \\
          	&q(\boldsymbol{\mu}_k|\boldsymbol{\Sigma}_k) = \mathcal{N}(\boldsymbol{\mu}_k|\tilde{\boldsymbol{\mu}}_k,\tilde{\eta}_k^{-1}\boldsymbol{\Sigma}_k)~, \\
          &q(\boldsymbol{\Sigma}_k) = \mathcal{IW}(\boldsymbol{\Sigma}_k|\tilde{\gamma}_k,\tilde{\boldsymbol{\Sigma}}_k)~.
          \end{aligned}
          \right .
      \end{equation*}
      
      \begin{flushleft}
      During the M-step, update rules for hyper-parameters are 
      \end{flushleft}

 	\begin{align*}
 					&\tilde{k}_k=k_0+N\bar{\pi}_k~, \\
					&\tilde{\eta}_k=\eta_0+N\bar{\omega}_k~, \\
					&\tilde{\boldsymbol{\mu}}_k=\frac{\eta_0\boldsymbol{\mu}_0+N\bar{\omega}_k\boldsymbol{\mu}^x_k}{\tilde{\eta}_k}~, \\
					&\tilde{\gamma}_k=\gamma_0+N\bar{\pi}_k~, \\
 					&\tilde{\boldsymbol{\Sigma}}_k=\boldsymbol{\Sigma}_0+\frac{N\bar{\omega}_k\eta_0}{\tilde{\eta}_k}\left(\boldsymbol{\mu}^x_k-\boldsymbol{\mu}_0\right)\left(\boldsymbol{\mu}^x_k-\boldsymbol{\mu}_0\right)^T+\boldsymbol{\Sigma}^x_{k}\; ,
 	\end{align*}
	\begin{flushleft}
	where  auxiliary variables are obtained as follows
	\end{flushleft}			
	\begin{align*}
				&\bar{\pi}_k=\frac{1}{N}\sum\limits_n\mathbb{E}_{\mathbf{U},\mathbf{Z}}[z_{n,k}]~,\\
				&\bar{\omega}_k=\frac{1}{N}\sum_n\mathbb{E}_{\mathbf{U},\mathbf{Z}}[z_{n,k}]\mathbb{E}_{\mathbf{U},\mathbf{Z}}[u_{n,k}]~,\\
				 &\boldsymbol{\mu}^x_k=\frac{1}{N\bar{\omega}_k}\sum_n\mathbb{E}_{\mathbf{U},\mathbf{Z}}[z_{n,k}]\mathbb{E}_{\mathbf{U},\mathbf{Z}}[u_{n,k}]\boldsymbol{x}_n~,\\
				 &\boldsymbol{\Sigma}^x_k=\frac{1}{N\bar{\omega}_k}\sum_n\mathbb{E}_{\mathbf{U},\mathbf{Z}}[z_{n,k}]\mathbb{E}_{\mathbf{U},\mathbf{Z}}[u_{n,k}]\left(\boldsymbol{x}_n-\boldsymbol{\mu}^x_k\right)\left(\boldsymbol{x}_n-\boldsymbol{\mu}^x_k\right)^T_.
	\end{align*}
                
Using the properties of the Dirichlet and the Inverse Wishart distribution, the following expectations are defined 
 \begin{align*}
 &\mathbb{E}_{\boldsymbol{\Theta}}[\log a_k]=\psi(\tilde{\kappa}_k)-\psi\left(\sum_{k'=1}^K\tilde{\kappa}_{k'}\right)~,\\
 &\mathbb{E}_{\boldsymbol{\Theta}}[\boldsymbol{\Sigma}_k^{-1}]=\tilde{\gamma}_k\tilde{\boldsymbol{\Sigma}}_k^{-1}~,\\
 &\mathbb{E}_{\boldsymbol{\Theta}}[\log \boldsymbol{|\Sigma}_k|]=\log|\tilde{\boldsymbol{\Sigma}}_k|-\sum_{i=1}^d\psi \left(\frac{\tilde{\gamma}_k+1-i}{2}\right)-d\log2~,\\
&\mathbb{E}_{\boldsymbol{\Theta}}[\mathcal{D}(\mathbf{x}_n,\boldsymbol{\mu}_k,\boldsymbol{\Sigma}_k)] =\tilde{\gamma}_k(\mathbf{x}_n-\tilde{\boldsymbol{\mu}}_k)^T\tilde{\boldsymbol{\Sigma}}_k^{-1}(\mathbf{x}_n-\tilde{\boldsymbol{\mu}}_k)+\frac{d}{\tilde{\eta}_k}\; .
			 \end{align*}


   \subsection{Variational posterior distributions for hyper-parameters}
	The assumption of independence between $\boldsymbol{\alpha}$ and $\boldsymbol{\beta}$ involves that two steps are required for the calculation of their independent posteriors distributions. Furthermore, since a conjugate prior has been designed in (\ref{indprior}), conjugate posterior distributions are obtained from the $\alpha$-step and the $\beta$-step such that :
    
    \begin{align*}
    &q(\boldsymbol{\beta}|K)=\prod_{k=1}^Kq(\beta_k)~,\\
    &q(\boldsymbol{\alpha}|K)=\prod_{k=1}^Kq(\alpha_k)~,
    \end{align*}
         
         \begin{flushleft}
         	where :
         \end{flushleft}
         
    \begin{align}\label{postbeta}
    	&q(\beta_k)=\mathcal{G}(\beta_k|\tilde{s}_k,\tilde{q}_k)\;,\\ \label{postalpha}
        &q(\alpha_k)=\frac{1}{M_k}\frac{\tilde{p}_k^{\alpha_k-1}}{\Gamma(\alpha_k)^{\tilde{r}_k}}\mathbb{I}_{\{\alpha_k>0\}}
    \end{align}
    
    \begin{flushleft}
    with
    \end{flushleft}
    
    \begin{equation*}
    	M_k=\int\frac{\tilde{p}_k^{\alpha_k-1}}{\Gamma(\alpha_k)^{\tilde{r}_k}}\mathbb{I}_{\{\alpha_k>0\}}d\alpha_k\;.
    \end{equation*}
    
   	Parameters $\tilde{p}_k,\tilde{q}_k, \tilde{r}_k$ and $\tilde{s}_k$ are updated as follows :
    
    \begin{align*}
	&\tilde{p}_k=p_0\exp \left(N\bar{\delta}_k+N\bar{\pi}_k\mathbb{E}_{\boldsymbol{\beta}}[\log\beta_k]\right)~,\\
        &\tilde{q}_k=q_0+N\bar{\omega}_k~,\\
    	&\tilde{r}_k=r_0+N\bar{\pi}_k~,\\
        &\tilde{s}_k=s_0+N\bar{\pi}_k\mathbb{E}_{\boldsymbol{\alpha}}[\alpha_k]~.
    \end{align*}
    
    \begin{flushleft}
    where 
    \end{flushleft}
    
    \begin{equation*}
    \bar{\delta}_k=\frac{1}{N}\sum_n\mathbb{E}_{\mathbf{U},\mathbf{Z}}[z_{nk}]\mathbb{E}_{\mathbf{U},\mathbf{Z}}[\log u_{nk}]~.
    \end{equation*}    
 
\section{Expectations in lower bound}
\label{sec:expectations}

The lower bound (\ref{bound}) is  proven to increase at each VB iteration and its difference between two iterations can be used as a stop criterion. The introduction of $(\boldsymbol{\alpha},\boldsymbol{\beta})$ slightly modifies the lower bound since the prior distribution (\ref{indprior}) as well as the posterior distributions (\ref{postbeta}) and (\ref{postalpha}) have to be taken into account. Lower bound elements related to $(\boldsymbol{\alpha},\boldsymbol{\beta})$ are presented below, others can be found in the Appendix B.\\

    Modifications related to $\mathbb{E}_q\left[\log p(\mathbf{X},\mathbf{H}|K)\right]$ are :
	\begin{align*}
    &\mathbb{E}_q[\log p(\mathbf{U}|\mathbf{Z},\boldsymbol{\Theta,\alpha,\beta},K)]=\sum\limits_{n,k}\mathbb{E}_{\mathbf{Z}}[z_{n,k}]\\
    &\bigg(\mathbb{E}_{\boldsymbol{\alpha}}[\alpha_k]\mathbb{E}_{\boldsymbol{\beta}}[\log \beta_k]-\mathbb{E}_{\boldsymbol{\alpha}}[\log\Gamma (\alpha_k)]\\
    &+(\mathbb{E}_{\boldsymbol{\alpha}}[\alpha_k]-1)\mathbb{E}_{\mathbf{U}}[\log u_{n,k}]-\mathbb{E}_{\boldsymbol{\beta}}[\beta_k]\mathbb{E}_{\mathbf{U}}[u_{n,k}]\bigg)
    \end{align*}
	\begin{flushleft}
	and
	\end{flushleft}
	
	\begin{align*}
				&\mathbb{E}_q[\log p(\boldsymbol{\alpha},\boldsymbol{\beta}|K)]=\sum\limits_k-\log M_k^0+(\mathbb{E}_{\boldsymbol{\alpha}}[\alpha_k]-1)\ln p_0\\
                &-r_0\mathbb{E}_{\boldsymbol{\alpha}}[\log\Gamma(\alpha_k)]+s_0\log q_0 -\log\Gamma(s_0)\\
                &+(s_0-1)\mathbb{E}_{\boldsymbol{\beta}}[\log\beta_k]-q_0\mathbb{E}_{\boldsymbol{\beta}}[\beta_k]~.
    	\end{align*}
             
	Modifications related to  $\mathbb{E}_q\left[\log q(\mathbf{H})\right]$ are :
    \begin{align*}
    &\mathbb{E}_q[\log q(\mathbf{U}|\mathbf{Z},\boldsymbol{\Theta,\alpha,\beta},K)]=\sum\limits_{n,k}\mathbb{E}_{\mathbf{Z}}[z_{nk}]\bigg(\tilde{\alpha}_k\log \tilde{\beta}_k-\log\Gamma (\tilde{\alpha}_k)\\
    &+(\tilde{\alpha}_k-1)\mathbb{E}_{\mathbf{U}}[\log u_{nk}]-\tilde{\beta}_k\mathbb{E}_{\mathbf{U}}[u_{n,k}]\bigg)
    \end{align*}
    \begin{flushleft}
	and
	\end{flushleft}
	
    \begin{align*}
				&\mathbb{E}_q[\log q(\boldsymbol{\alpha},\boldsymbol{\beta}|K)]=\sum\limits_k-\log M_k+(\mathbb{E}_{\boldsymbol{\alpha}}[\alpha_k]-1)\log \tilde{p}_k\\
                &-\tilde{r}_k\mathbb{E}_{\boldsymbol{\alpha}}[\log\Gamma(\alpha_k)]+\tilde{s}_k\log \tilde{q}_k-\log\Gamma(\tilde{s}_k)+(\tilde{s}_k-1)\mathbb{E}_{\boldsymbol{\beta}}[\log \beta_{k}]\\
                &-\tilde{q}_k\mathbb{E}_{\boldsymbol{\beta}}[\beta_{k}]~.
   \end{align*}
 Posterior expectations of $\beta_k$ are derived from the posterior Gamma distribution (\ref{postbeta}) properties and can easily be computed by
     \begin{align*}
    &\mathbb{E}_{\boldsymbol{\beta}}[\beta_k]=\frac{\tilde{s}_k}{\tilde{q}_k}~,\\
    & \mathbb{E}_{\boldsymbol{\beta}}[\log \beta_k]=\psi(\tilde{s}_k)-\log \tilde{q}_k~\; .
    \end{align*}
    
    However, expectations depending on $\alpha_k$ are intractable     
    \begin{align}\label{exp1}
    &\mathbb{E}_{\boldsymbol{\alpha}}[\alpha_k]=\int\alpha_kp(\alpha_k|\tilde{p}_k,\tilde{r}_k)d\alpha_k~,\\ \label{exp2}
   &\mathbb{E}_{\boldsymbol{\alpha}}[\log\Gamma(\alpha_k)]=\int\log\Gamma(\alpha_k)p(\alpha_k|\tilde{p}_k,\tilde{r}_k)d\alpha_k~\; .
    \end{align}
    
    Since lower bound calculation is required as a stop criterion, expectations (\ref{exp1}) and (\ref{exp2}) have to be approximated. A deterministic method \cite{Tierney1986} based on Laplace approximation is then applied. 
    
    \subsection{Laplace approximation}
    That method consists in evaluating the posterior moments  and variances of a positive function $g(\alpha_k)$ as follows : 
    \vspace{-10pt}
    
    \begin{equation*}\label{approx}
    	\hat{\mathbb{E}}_n[g(\alpha_k)]=\left(\frac{\sigma_1(\hat{\alpha}_k^1)}{\sigma_2(\hat{\alpha}_k^2)}\right)^{\frac{1}{2}}\exp \left(-n\left(l_1(\hat{\alpha}_k^1)-l_2(\hat{\alpha}_k^2)\right)\right)\;,
    \end{equation*}
    
    \begin{flushleft}
    where  $n\in\mathbb{N}^*$, $\hat{\alpha}_k^i$ is the minimizer for $l_i(\alpha_k)$, $\sigma_i(\alpha_k)$ is the inverse of the Hessian  of $l_i(\alpha_k)$ and :
    \end{flushleft}
    \vspace{-10pt}
    
    \begin{equation*}\label{ldef}
          \left \{
          \begin{aligned}
          	&l_1(\alpha_k)=-n^{-1}\ln\left(g(\alpha_k)p(\alpha_k|\tilde{p}_k,\tilde{q}_k,\tilde{r}_k,\tilde{s}_k)\right) \;,\\
          &l_2(\alpha_k)=-n^{-1}\ln\left(p(\alpha_k|\tilde{p}_k,\tilde{q}_k,\tilde{r}_k,\tilde{s}_k)\right)~.
          \end{aligned}
          \right.
     \end{equation*}
   However,  the positivity of the function $\log\Gamma(\alpha_k)$ is not always verified for any $\alpha_k>0$. In the negative case, an importance sampling method  \cite{Hastings1970} is applied to evaluate  $\mathbb{E}_{\boldsymbol{\alpha}}[\log\Gamma(\alpha_k)]$. 
   
     \begin{table}[]
\centering
\caption{Code parameters}
\label{data_code}
\begin{tabular}{ll}
\hline
\hline
Code name     & Code parameters \\ \hline
Linear        & B=250 MHz      \\
Quadratic     & B=250 MHz      \\
Costas        & M=7, B=250 MHz    \\
Barker        & Code 13         \\
Frank, P1, P2 & M=64            \\
Zadoff        & M=64, r=7, q=32 \\
P3, P4        & M=64            \\ \hline \hline
\end{tabular}
\end{table}

   \subsection{Importance sampling}
   For any distribution $p$ and any measurable function $g$, the importance sampling approximates $\mathbb{E}_p[g(x)]=\int g(x)p(x)dx$ by 
     \vspace{-10pt}
     
     \begin{equation}\label{importance}	
     \hat{\mathbb{E}}_n[g(\mathbf{x})]=\sum\limits_{j=1}^J g(x_j)\frac{p(x_j)}{q(x_j)}
     \end{equation}
     
     \begin{flushleft}\label{support}
      	where  $\mathbf{x}=\left(x_1,\ldots,x_J\right)$ is sampled from an instrumental distribution $q$ satisfying supp$(p)$ $\subset$ supp$(q)$ .
     \end{flushleft}
     Then, $\hat{\mathbb{E}}_{n}[\log\Gamma(\alpha_k)]$ is computed by choosing $p$ as the posterior distribution obtained in (\ref{postalpha}) and $q$  as a Gamma distribution whose parameters are designed to ensure a finite variance for (\ref{importance}).
     It can be noted that importance sampling approach can also be used in the positive case but due to its higher computational cost Laplace approximation is preferred.

\section{Experiments}
\label{sec:experiments}

\begin{figure}
\centering
\includegraphics[scale=0.25]{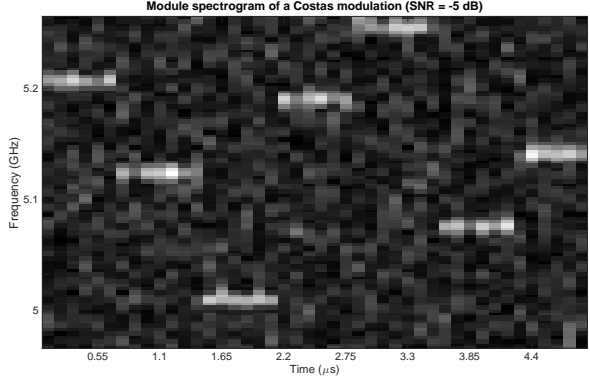}
\caption{Module spectrogram of a Costas signal (SNR = -5 dB)}
\label{Costas_spec}
\end{figure}

In this section, the proposed method is performed on a set of simulated data.  For comparison, the Bayesian Student's t-mixture model  from \cite{Archambeau2007} is also evaluated. 
Two experiments are  carried out  to evaluate  classification and clustering performances with respect to a range of signal-to-noise ratios. First, simulated data  are introduced and prepocessing techniques are detailed. Then, both experiments are described with their associated error measures. Finally, performances are shown to exhibit the effectiveness of the proposed model.

\subsection{Data}
LPI signals are simulated with respect to equations (\ref{eqsignal})  and (\ref{pulse_equation}). The reference frequency $f_0$, the pulse duration $T$ and the sampling frequency are respectively chosen as 5 GHz, 5 $\mu$s and 1 GHz. For each type of modulation, the complex envelope $u(t)$ in (\ref{pulse_equation}) is chosen  according to each modulation definition reported in subsections \ref{FM} and \ref{PM}. Parameters of LPI signals are available in Table \ref{data_code}. The variance $\sigma_n^2$ of the noise component $n(t)$ in  (\ref{eqsignal})  is taken with respect to the SNR defined  in \cite{Ravi2017} as 
\begin{equation*}
\text{SNR(dB)}=10\times \log_{10}\left(\frac{\sigma^2_s}{\sigma_n^2}\right)
\end{equation*}
where $\sigma^2_s$ is the variance of original input signal. Since we are interested in ESM applications where received signals are embedded in noise, only SNRs from -15dB to 0dB are considered. Then for each SNR value, 1000 simulations of each LPI signal are carried out and a Short Term Fourier Transform (STFT) \cite{Allen1977} is applied on them to emphasise their time-frequency features. At last a Principal Component Analysis \cite{Jolliffe1986} is performed on modules of the STFTs to reduce the dimension of data and only the first components concentrating variance of the modules are kept. Figure \ref{Costas_spec} shows the module spectrogram of a Costas signal with SNR of -5 dB.


\subsection{Experiments}
 The classification experiment tests the ability of each algorithm to assign any data to their true classes knowing the clusters number $K$ whereas the clustering one aims to determine the ability of each algorithm to restore the true clusters according to an a priori number of clusters $L0>K$. During each experiment, $M$ simulations are  performed. For each simulation $m\in\{1,\ldots,M\}$, an unique random  non supervised initialization  is retained  for both algorithms. It consists in drawing  responsibilities $r_{nk}$ from a uniform distribution on $[0,1]$ and  normalizing  as in (\ref{norm}) and sampling center $\boldsymbol{\mu}_k$ from a multivariate normal distribution whose mean vector, respectively covariance matrix, is the mean, respectively the covariance matrix, of  the observed  data  $\mathbf{X}$. Clusters covariance matrices are initialized from identity matrices. Both algorithms terminate when the difference of log-likelihood bound is less than $10^{-8}$. For the classification experiment, respectively the clustering experiment, the Accuracy measure (Acc) and the Cross-Entropy Loss (CEL), respectively the Adjusted Rand Index (ARI) \cite{Hubert1985}, are computed from obtained results to compare algorithms performances.

%
 
 \subsection{Results}
\begin{table}[]
\centering
\caption{Average Accuracy rate (Acc) and Cross-Entropy Loss (CEL) comparing the proposed GSMM to the SMM (M=100)}
\label{Acc}
\begin{tabular}{ccccc}
\hline
\hline
\multicolumn{1}{l}{} & \multicolumn{2}{c}{Acc}    & \multicolumn{2}{c}{CEL}    \\ 
SNR                  & SMM      & GSMM            & SMM      & GSMM            \\ \hline 
\multirow{2}{*}{-15} & 0.1517   & \textbf{0.5897} & 2.2340   & \textbf{0.6361} \\
                     & (0.0605) & (0.0768)        & (0.0211) & (0.5328)        \\
\multirow{2}{*}{-10} & 0.6703   & \textbf{0.7090} & 2.0257   & \textbf{1.0071} \\
                     & (0.1173) & (0.0953)        & (2.7106) & (1.2453)        \\
\multirow{2}{*}{-5}  & 0.6944   & \textbf{0.7106} & 1.9419   & \textbf{0.9621} \\
                     & (0.1329) & (0.1292)        & (2.5713) & (1.3276)        \\
\multirow{2}{*}{0}   & 0.7088   & \textbf{0.7158} & 2.3030   & \textbf{1,1528} \\
                     & (0.1213) & (0.1224)        & (2.8299) & (1.2674)        \\ \hline \hline
\multicolumn{5}{l}{The numbers in parenthesis are the standard deviations}     \\
\multicolumn{5}{l}{of the corresponding quantities.}                          
\end{tabular}
\end{table}
 The mean and standard deviation of Accuracy rate and Cross-Entropy Loss are shown in Table \ref{Acc}. The proposed method  obtains more accurate classification performances with lower variance  since the GSMM presents a higher Accuracy rate and a lower Cross-Entropy Loss than  the standard method for all SNR values. Regarding the lowest SNR (-15dB), the GSMM outperforms the SMM meaning that the GSMM is less sensitive to initialization even in the presence of noise. The GSMM classification results with -10dB SNR  are more precisely  detailed in an average confusion matrix  in Table \ref{CM}. The GSMM successfully separated frequency modulations from phase modulations but could not exactly tell the difference  between some phase modulations such as P2 and P4. That lack of differentiation can be explained either by the choice of identical duration parameters $t_b$ for all phase modulations or by the hard dimension reduction applied on data since only the six first components were kept over five thousand components. Both of these reasons can  create non separable data that mixture models can not handle. Results of the clustering experiment are also detailed in Table \ref{ARI} where the GSMM  shows higher ARI with lower variance than the SMM for both a priori numbers of clusters  $L0$ . Therefore these experiments suggest that the GSMM is less sensitive to initialization and produce more accurate results  by allowing  a less restrictive  modeling of data.


\begin{table}[]
\centering
\caption{Average confusion matrix for the GSMM results with SNR of $-10\; dB$}
\label{CM}
\centering
\resizebox{\columnwidth}{!}{%
\begin{tabular}{ccccccccccc}
\hline \hline
       & LFM           & QFM           & Costas         & Barker        & Frank         & P1            & P2            & Zadoff        & P3            & P4            \\ \hline
LFM    & \textbf{97\%} & -             & 3\%            & -             & -             & -             & -             & -             & -             & -             \\
QFM    & 2\%           & \textbf{97\%} & 1\%            & -             & -             & -             & -             & -             & -             & -             \\
Costas & -             & -             & \textbf{100\%} & -             & -             & -             & -             & -             & -             & -             \\
Barker & 1\%           & -             & 2\%            & \textbf{32\%} & 39\%          & -             & -             & 24\%          & 2\%           & -             \\
Frank  & 1\%           & -             & 2\%            & 11\%          & \textbf{68\%} & 2\%           & 1\%           & 12\%          & 3\%           & -             \\
P1     & 3\%           & 2\%           & 3\%            & 4\%           & 3\%           & \textbf{53\%} & -             & 17\%          & 12\%          & 3\%           \\
P2     & 1\%           & 1\%           & -              & -             & -             & -             & \textbf{13\%} & -             & 1\%           & 84\%          \\
Zadoff & 2\%           & 3\%           & 2\%            & 3\%           & -             & 1\%           & -             & \textbf{87\%} & 2\%           & -             \\
P3     & 2\%           & 2\%           & 3\%            & 1\%           & 1\%           & 1\%           & 1\%           & 1\%           & \textbf{88\%} & -             \\
P4     & 1\%           & 1\%           & -              & 1\%           & -             & -             & 12\%          & -             & 1\%           & \textbf{84\%} \\ \hline \hline
\end{tabular}}
\end{table}

\begin{table}[]
\centering
\caption{Average Adjusted Rand Index comparing the proposed GSMM to the SMM (M=100)}
\label{ARI}
\begin{tabular}{ccccc}
\hline
\hline
\multicolumn{1}{l}{} & \multicolumn{2}{c}{L0=15}    & \multicolumn{2}{c}{L0=20}    \\ 
SNR                  & SMM      & GSMM            & SMM      & GSMM            \\ \hline 
\multirow{2}{*}{-15} & 0.3934   & \textbf{0.3999} & 0.4078   & \textbf{0.4083} \\
                     & (0.0614) & (0.0694)        & (0.0554) & (0.546)        \\
\multirow{2}{*}{-10} & 0.7071   & \textbf{0.7243} & 0.7256   & \textbf{0.7337} \\
                     & (0.0798) & (0.0704)        & (0.0690) & (0.0561)        \\
\multirow{2}{*}{-5}  & 0.7914   & \textbf{0.8099} & 0.8029   & \textbf{0.8122} \\
                     & (0.0752) & (0.0641)        & (0.0661) & (0.0530)        \\
\multirow{2}{*}{0}   & 0.8061   & \textbf{0.8235} & 0.8061   & \textbf{0.8152} \\
                     & (0.0680) & (0.0574)        & (0.0655) & (0.0537)        \\ \hline \hline
\multicolumn{5}{l}{The numbers in parenthesis are the standard deviations}     \\
\multicolumn{5}{l}{of the corresponding quantities.}                          
\end{tabular}
\end{table}

\section{CONCLUSION}
\label{sec:conclusion}    

In this paper, we develop an unsupervised mixture model to classify and cluster Low Probability of Intercept (LPI) radar signals. LPI radar signals are particularly designed to be embedded in noise, hence a generalized multivariate Student-t mixture model, known for its robustness to outliers, is chosen. Thanks to the introduction of a novel prior distribution for its hyper-parameters, the generalized model can handle sensitivity of mixture models to initialization. Model learning is processed through a Variational Bayes inference where approximation methods are applied to estimate intractable expectations of the hyper-parameters posterior distribution. Experiments on various simulated data showed that the proposed approach is less sensitive to initialization and can outperform the standard model in classification and clustering tasks.

\section*{Appendices}

\subsection*{A : Proof}

\begin{flushleft}
For any $p,r>0$, the integral of interest is 
\end{flushleft}

\begin{equation*}
	M=\int_0^\infty \frac{p^{x-1}}{\Gamma(x)^r}dx
\end{equation*}

That can be reformulated as 

\begin{align*}
	M&=\int_0^\infty \exp\left( \ln p(x-1)-r\ln \Gamma(x)\right)dx\\
    &=\int_0^\infty f(x|p,r)dx
\end{align*}

\begin{flushleft}
	where $f$ is a strictly positive function with a first  derivative equals to
\end{flushleft}
 
 \begin{equation*}
 f'(x|p,r)=\left(\ln p-r\psi(x)\right)f(x|p,r)
 \end{equation*}
 
 The function $x\rightarrow\ln p-r\psi(x)$ is strictly decreasing on $\mathbb{R}^*_+$ and intersects the x-axis at $x_0=\psi^{-1}(\frac{\ln p}{r})$. Then, $f'$ is strictly positive on $]0,x_0[$, respectively strictly negative on $]x_0,+\infty[$, that leads $f$ is strictly increasing on $]0,x_0[$, respectively strictly decreasing on $]x_0,+\infty[$. Therefore, $f$ admits a maximum value at $x_0$ and is upper bounded.
 Limits of $f$ has to be calculated to determine the existence of a lower bound. 
 
  \begin{equation*}
 	\lim\limits_{\substack{x \rightarrow 0 \\ x>0}} f(x)=\lim\limits_{\substack{x \rightarrow 0 \\ x>0}} \exp (-r\Gamma(x))=0
 \end{equation*}
 is obtained using the gamma function properties.
 However,
 \begin{equation*}
 	\lim\limits_{x\rightarrow\infty}f(x)=\lim\limits_{x\rightarrow\infty}\exp\left( \ln p(x-1)-r\ln \Gamma(x)\right)=0
 \end{equation*}
 is not always satisfied for any $p$ and $r$. If $p\le 1$ and $r\ge 0$, the limit is verified.
 In that particular case, $f$ is lower bounded by $0$ and upper bounded by $f(x_0)$ and that involves $M$ is proper and converges.

\subsection*{B : Lower bound elements}

\begin{flushleft}
    Elements related to $\mathbb{E}_q\left[\log p(\mathbf{X},\mathbf{H}|K)\right]$ are :
    \end{flushleft}
    
    \begin{align*}
    &\mathbb{E}_q[\log p(\mathbf{X}|\mathbf{U,Z},\boldsymbol{\Theta},K)]=\sum\limits_{n,k}\mathbb{E}_{\mathbf{U,Z}}[z_{nk}]\bigg\{\\
				&-\frac{\mathbb{E}_{\boldsymbol{\Theta}}[\log |\boldsymbol{\Sigma}_k|]}{2}-\frac{d}{2}(\log 2\pi-\mathbb{E}_{\mathbf{U,Z}}[\log u_{nk}])\\
				&-\frac{\mathbb{E}_{\mathbf{U,Z}}[u_{nk}]}{2}\mathbb{E}_{\boldsymbol{\Theta}}\left[\mathcal{D}(\mathbf{x}_n,\boldsymbol{\mu}_k,\boldsymbol{\Sigma}_k)\right]\bigg\}
    \end{align*}
    
    \begin{flushleft}
    and
    \end{flushleft}
    
    \vspace{-12pt}
    
    \begin{equation*}
    \mathbb{E}_q[\log p(\mathbf{Z}|\boldsymbol{\Theta},K)]=\sum\limits_{n,k}\mathbb{E}_{\mathbf{U,Z}}[z_{nk}]\mathbb{E}_{\boldsymbol{\Theta}}[\log a_{k}]
    \end{equation*}
    
     \begin{flushleft}
    and
    \end{flushleft}
    
    \vspace{-12pt}
    
    \begin{equation*}
    	\mathbb{E}_q[\log p(\mathbf{a}|K)]= \log c_{\mathcal{D}}(\kappa_0)+\sum\limits_k(\kappa_0-1)\mathbb{E}_{\boldsymbol{\Theta}}[\log a_k]
    \end{equation*}
    
     \begin{flushleft}
    and
    \end{flushleft}
    
    \vspace{-12pt}
    
    \begin{align*}
    &\mathbb{E}_q[\log p(\boldsymbol{\mu}|\boldsymbol{\Sigma},K)]=\sum\limits_{k}-\frac{d}{2}(\log 2\pi-\log \eta_0)-\frac{\mathbb{E}_{\boldsymbol{\Theta}}[\log |\boldsymbol{\Sigma}_k|]}{2}\\
    &-\frac{\eta_0}{2}\mathbb{E}_{\boldsymbol{\Theta}}\left[\mathcal{D}(\boldsymbol{mu}_k,\boldsymbol{\mu}_0,\boldsymbol{\Sigma}_k)\right]
    \end{align*}
    
    \begin{flushleft}
    and
    \end{flushleft}
    
    \vspace{-12pt}
    
\begin{align*}				
				&\mathbb{E}_q[\log p(\boldsymbol{\Sigma}|K)]=\sum\limits_k\log c_{\mathcal{IW}}(\gamma_0,\boldsymbol{\Sigma}_0)-\frac{tr\{\boldsymbol{\Sigma}_0\mathbb{E}_{\boldsymbol{\Theta}}[\boldsymbol{\Sigma}_k^{-1}]\}}{2}\\
				&-\frac{\gamma_0+d+1}{2}\mathbb{E}_{\boldsymbol{\Theta}}[\log |\boldsymbol{\Sigma}_k|]\;.
\end{align*}

\begin{flushleft}
Elements related to  $\mathbb{E}_q\left[\log q(\mathbf{H})\right]$ are :
\end{flushleft}

\begin{equation*}
    \mathbb{E}_q[\log q(\mathbf{Z}|\boldsymbol{\Theta},K)]=\sum\limits_{n,k}\mathbb{E}_{\mathbf{U,Z}}[z_{nk}]\log r_{nk}
    \end{equation*}
    
     \begin{flushleft}
    and
    \end{flushleft}
    
    \vspace{-12pt}
    
    \begin{equation*}
    	\mathbb{E}_q[\log q(\mathbf{a}|K)]= \log c_{\mathcal{D}}(\tilde{\boldsymbol{\kappa}})+\sum\limits_k(\tilde{\kappa}_k-1)\mathbb{E}_{\boldsymbol{\Theta}}[\log a_k]
    \end{equation*}
    
     \begin{flushleft}
    and
    \end{flushleft}
    
    \vspace{-12pt}
    
    \begin{equation*}
    \mathbb{E}_q[\log q(\boldsymbol{\mu}|\boldsymbol{\Sigma},K)]=\sum\limits_{k}-\frac{d}{2}(\log 2\pi-\log \tilde{\eta}_k)-\frac{\mathbb{E}_{\boldsymbol{\Theta}}[\log |\boldsymbol{\Sigma}_k|]}{2}
    \end{equation*}
    
    \begin{flushleft}
    and
    \end{flushleft}
    
    \vspace{-12pt}
    
\begin{align*}				
				&\mathbb{E}_q[\log q(\boldsymbol{\Sigma}|K)]=\sum\limits_k\log c_{\mathcal{IW}}(\tilde{\gamma_k},\tilde{\boldsymbol{\Sigma}}_k)-\frac{tr\{\tilde{\boldsymbol{\Sigma}}_k\mathbb{E}_{\boldsymbol{\Theta}}[\boldsymbol{\Sigma}_k^{-1}]\}}{2}\\
				&-\frac{\tilde{\gamma}_k+d+1}{2}\mathbb{E}_{\boldsymbol{\Theta}}[\log |\boldsymbol{\Sigma}_k|]\;.
\end{align*}






%

%
%

\bibliographystyle{plain}
\bibliography{refs}

\begin{thebibliography}{10}

\bibitem{Allen1977}
J.~Allen.
\newblock Short term spectral analysis, synthesis, and modification by discrete
  {F}ourier transform.
\newblock {\em IEEE Transactions on Acoustics, Speech, and Signal Processing},
  25(3):235--238, Jun 1977.

\bibitem{Archambeau2007}
C{\'e}dric Archambeau and Michel Verleysen.
\newblock Robust {B}ayesian clustering.
\newblock {\em Neural Networks}, 20(1):129--138, 2007.

\bibitem{Barker1953}
RH~Barker.
\newblock Group syncronization of binary digital systems.
\newblock {\em Communication theory}, pages 273--287, 1953.

\bibitem{Copeland2002}
D.~B. Copeland and P.~E. Pace.
\newblock Detection and analysis of {FMCW} and {P}-4 polyphase {LPI} waveforms
  using quadrature mirror filter trees.
\newblock In {\em 2002 IEEE International Conference on Acoustics, Speech, and
  Signal Processing}, volume~4, pages IV--3960--IV--3963, May 2002.

\bibitem{Costas1984}
John~P Costas.
\newblock A study of a class of detection waveforms having nearly ideal range.
  {D}oppler ambiguity properties.
\newblock {\em Proceedings of the IEEE}, 72(8):996--1009, 1984.

\bibitem{Dempster1977}
Arthur~P. Dempster, Nan~M. Laird, and Donald~B. Rubin.
\newblock Maximum likelihood from incomplete data via the {EM} algorithm.
\newblock {\em Journal of the royal statistical society. Series B
  (methodological)}, pages 1--38, 1977.

\bibitem{Frank1962}
R~Frank, S~Zadoff, and R~Heimiller.
\newblock Phase shift pulse codes with good periodic correlation properties.
\newblock {\em IRE Transactions on Information Theory}, 8(6):381--382, 1962.

\bibitem{Gencol2016}
K.~Gencol.
\newblock A set of features for classification of intrapulse modulations.
\newblock In {\em 2016 24th Signal Processing and Communication Application
  Conference (SIU)}, pages 2113--2116, May 2016.

\bibitem{Hastings1970}
W.~Keith Hastings.
\newblock Monte {C}arlo sampling methods using {M}arkov chains and their
  applications.
\newblock {\em Biometrika}, 57(1):97--109, 1970.

\bibitem{Hubert1985}
Lawrence Hubert and Phipps Arabie.
\newblock Comparing partitions.
\newblock {\em Journal of classification}, 2(1):193--218, 1985.

\bibitem{Jolliffe1986}
Ian~T Jolliffe.
\newblock Principal component analysis and factor analysis.
\newblock In {\em Principal component analysis}, pages 115--128. Springer,
  1986.

\bibitem{Jordan1994}
Michael~I Jordan and Robert~A Jacobs.
\newblock Hierarchical mixtures of experts and the {EM} algorithm.
\newblock {\em Neural computation}, 6(2):181--214, 1994.

\bibitem{Ravi2017}
T.~Ravi Kishore and K.~D. Rao.
\newblock Automatic intrapulse modulation classification of advanced {LPI}
  radar waveforms.
\newblock {\em IEEE Transactions on Aerospace and Electronic Systems},
  53(2):901--914, April 2017.

\bibitem{Levanon2004}
Nadav Levanon and Eli Mozeson.
\newblock {\em Radar signals}.
\newblock John Wiley \& Sons, 2004.

\bibitem{Milne2002}
P.~R. Milne and P.~E. Pace.
\newblock Wigner distribution detection and analysis of {FMCW} and {P}-4
  polyphase {LPI} waveforms.
\newblock In {\em 2002 IEEE International Conference on Acoustics, Speech, and
  Signal Processing}, volume~4, pages IV--3944--IV--3947, May 2002.

\bibitem{Nguyen2014}
T.~M. Nguyen and Q.~M.~J. Wu.
\newblock Bounded asymmetrical {S}tudent's-t mixture model.
\newblock {\em IEEE Transactions on Cybernetics}, 44(6):857--869, June 2014.

\bibitem{Opper2001}
Manfred Opper and David Saad.
\newblock {\em Advanced mean field methods: Theory and practice}.
\newblock MIT press, 2001.

\bibitem{Peel2000}
David Peel and Geoffrey~J. McLachlan.
\newblock Robust mixture modelling using the t distribution.
\newblock {\em Statistics and computing}, 10(4):339--348, 2000.

\bibitem{Quandt1978}
Richard~E. Quandt and James~B. Ramsey.
\newblock Estimating mixtures of normal distributions and switching
  regressions.
\newblock {\em Journal of the American statistical Association},
  73(364):730--738, 1978.

\bibitem{Schleher1986}
D.~Curtis Schleher.
\newblock Introduction to {E}lectronic {W}arfare.
\newblock Technical report, Eaton Corp., AIL Div., Deer Park, NY, 1986.

\bibitem{Sun2017}
J.~Sun, A.~Zhou, S.~Keates, and S.~Liao.
\newblock Simultaneous {B}ayesian clustering and feature selection through
  {S}tudent's t mixtures model.
\newblock {\em IEEE Transactions on Neural Networks and Learning Systems},
  PP(99):1--13, 2017.

\bibitem{Svensen2005}
Markus Svens{\'e}n and Christopher~M. Bishop.
\newblock Robust {B}ayesian mixture modelling.
\newblock {\em Neurocomputing}, 64:235--252, 2005.

\bibitem{Tierney1986}
Luke Tierney and Joseph~B. Kadane.
\newblock Accurate approximations for posterior moments and marginal densities.
\newblock {\em Journal of the {A}merican statistical association},
  81(393):82--86, 1986.

\bibitem{Turyn1963}
R~Turyn.
\newblock On {B}arker codes of even length.
\newblock {\em Proceedings of the IEEE}, 51(9):1256--1256, 1963.

\bibitem{Waterhouse1996}
Steve Waterhouse, David MacKay, Tony Robinson, et~al.
\newblock Bayesian methods for mixtures of experts.
\newblock {\em Advances in neural information processing systems}, pages
  351--357, 1996.

\bibitem{Wiley1982}
Richard~G Wiley.
\newblock Electronic {I}ntelligence: the analysis of radar signals.
\newblock {\em Dedham, MA, Artech House, Inc., 1982. 250 p}, 1982.

\bibitem{Zadoff1963}
Solomon Zadoff et~al.
\newblock Phase coded signal receiver, July~2 1963.
\newblock {U}{S} Patent 3,096,482.

\bibitem{Zhang2012}
H.~Zhang, Q.~M.~J. Wu, and T.~M. Nguyen.
\newblock Bayesian feature selection and model detection for {S}tudent's
  t-mixture distributions.
\newblock In {\em Proceedings of the 21st International Conference on Pattern
  Recognition (ICPR2012)}, pages 1631--1634, Nov 2012.

\bibitem{Zhang2013}
H.~Zhang, Q.~M.~J. Wu, and T.~M. Nguyen.
\newblock Image segmentation by a new weighted {S}tudent's t-mixture model.
\newblock {\em IET Image Processing}, 7(3):240--251, April 2013.

\end{thebibliography}

%








\end{document}